\documentclass[runningheads]{llncs}
\usepackage{graphicx}

\usepackage{tikz}
\usepackage{comment}
\usepackage{amsmath,amssymb} 
\usepackage{color}

\usepackage{caption} 
\usepackage{graphbox}  
\usepackage{mwe} 
\usepackage[export]{adjustbox} 
\usepackage{floatrow} 

\newcommand{\lcon}{L_{\text{content}}} 
\newcommand{\lsty}{L_{\text{style}}} 
\newcommand{\lmatch}{L_{\text{warp}}} 

\begin{document}
\pagestyle{headings}
\mainmatter
\def\ECCVSubNumber{5333}  

\title{Deformable Style Transfer} 

\titlerunning{Deformable Style Transfer}

\author{Sunnie S. Y. Kim\inst{1}\orcidID{0000-0002-8901-7233} \and
Nicholas Kolkin\inst{1}\orcidID{0000-0003-1233-1969} \and
Jason Salavon\inst{2}\orcidID{0000-0003-4885-9511} \and \\
Gregory Shakhnarovich\inst{1}\orcidID{0000-0003-4700-9398}
}

\authorrunning{S. Kim et al.}

\institute{\begin{tabular}{c@{\hskip 4em}c}%
             \inst{1}Toyota Technological Institute at %
             Chicago &\inst{2}University of Chicago\\ %
             \email{\{sunnie, nick.kolkin, greg\}@ttic.edu}
                     &\email{salavon@uchicago.edu}%
           \end{tabular}}

\maketitle

\begin{center}
\includegraphics[width=0.99\textwidth]{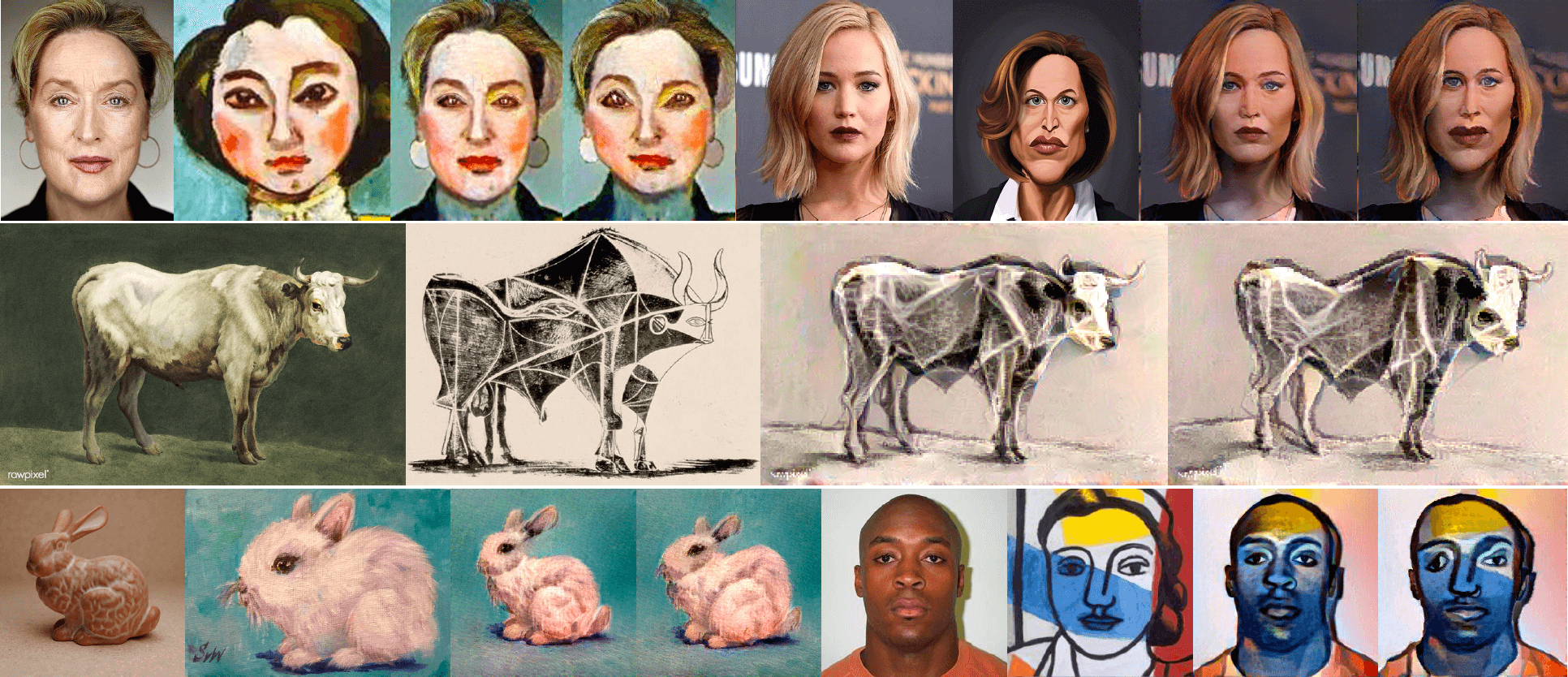}\\\captionof{figure}{Each set of four images contains (left to right) a content input, a style input, a standard style transfer output, and our proposed
 method's output. Sources of used images are available in the supplementary material.}\label{fig:fig1}
\end{center}

\begin{abstract}
Both geometry and texture are fundamental aspects of visual style. Existing style transfer methods, however, primarily focus on texture, almost entirely ignoring geometry. We propose deformable style transfer (DST), an optimization-based approach that jointly stylizes the texture and geometry of a content image to better match a style image. Unlike previous geometry-aware stylization methods, our approach is neither restricted to a particular domain (such as human faces), nor does it require training sets of matching style/content pairs. We demonstrate our method on a diverse set of content and style images including portraits, animals, objects, scenes, and paintings. Code has been made publicly available at \texttt{https://github.com/sunniesuhyoung/DST}.

\end{abstract}

\section{Introduction}\label{sec:intro}

The goal of style transfer algorithms is to re-render the content of one image using the style of one or several other images.  Most modern approaches~\cite{Gatys_2016_CVPR,Li_2016_CVPR,berger2016incorporating,gatys2017controlling,huang2017arbitrary,gu2018arbitrary,mechrez2018contextual,sanakoyeu2018styleaware,Kolkin_2019_CVPR} capture a definition of ``style'' that focuses on color and texture. Art historians and other experts on image creation, however, define style more broadly and almost always consider the shapes and geometric forms present in an artwork as integral parts of its style~\cite{hofstadter1983metamagical,elkins1996style}. Shape and form play a vital role in recognizing the distinctive style of many artists in painting (e.g. Picasso, Modigliani, El Greco), sculpture (e.g., Botero, Giacometti), and other forms of media. While the results of style transfer algorithms thus far have been impressive and have captured the public’s attention, we propose a method of extending style transfer to better match the geometry of an artist's style.

Style transfer methods that do not explicitly include geometry in their definition of style almost always keep the geometry of the content unchanged in the final output. This results in the outputs of these algorithms being easily identified as altered or ``filtered" versions of the content image, rather than novel images created using the content image as a reference. Our focus in this work is to integrate shape and geometry as important markers of style and loosen the constraints on content as a receptive canvas. We achieve this by introducing a domain-agnostic geometric deformation of the content image, optimized jointly with standard style transfer losses.

Our proposed method, deformable style transfer (DST), takes two images as the input: a content image and a style image. We assume these two images share a domain and have some approximate alignment (e.g. both are images of sitting animals). This is a general scenario likely to arise in recreational or artistic uses of style transfer, as well as in tasks such as data augmentation. The nature of this problem makes \emph{learning} to transfer style challenging since the variation in unconstrained domains and styles can not be reasonably presumed to be captured in any feasible training set. Therefore, like other style transfer work in this setting, we develop an optimization-based method, leveraging a pre-trained and fixed feature extractor derived from a convolutional network (CNN) trained for ImageNet classification.

There has been recent work on learning geometric style, using an explicit model of landmark constellations~\cite{Yaniv_2019_ACM} or a deformation model representing a specific style~\cite{Shi_2019_CVPR}. These methods require a collection of images in the chosen style, and work only in a specific domain (often faces, due to their importance in culture and applications). Hence, they are not applicable to our more general scenario. Nonetheless, we compare our results to those of the aforementioned methods in their specific domain of faces in Section~\ref{sec:results}, and find that our method produces equally aesthetically pleasing results despite it being more general.

In this work we propose the first, to our knowledge, method for incorporating geometry into one-shot, domain-agnostic style transfer. The key idea of DST is to find a smooth deformation, or spatial warping, of the content image that brings it into spatial alignment with the style image. This deformation is guided by a set of matching keypoints, chosen to maximize the feature similarity between paired keypoints of the two images. After roughly aligning the paired keypoints with a rigid rotation and scaling, a simple $\ell_2$ loss encourages warping our output image in such a way that the keypoints become spatially aligned. This deformation loss is regularized with a total variation penalty to reduce artifacts due to drastic deformations, and combined with the more traditional style and content loss terms. DST's joint, regularized objective simultaneously encourages preserving content, minimizing the style loss, and obtaining the desired deformation, weighing these goals against each other. This objective can be solved using standard iterative techniques.

To summarize the contributions of this work:
\begin{itemize}
\item We propose an optimization-based framework that endows style transfer algorithms with the explicit ability to deform a content image to match the geometry of a style image. Our flexible formulation also allows explicit user guidance and control of stylization tradeoffs.
\item We demonstrate, for the first time, geometry-aware style transfer in a one-shot scenario. In contrast to previous works that are limited to human faces, DST works for images in other domains, with the assumption that they share a domain and have some approximate alignment.
\item We evaluate DST on a range of style transfer instances, with images of faces, animals, vehicles, and landscapes, and through a user study demonstrate that our framework can augment existing style transfer algorithms to dramatically improve the perceived stylization quality, at minimal cost to the percieved content preservation.
\end{itemize}

\section{Related Work}\label{sec:related}

Early style transfer methods relied on hand-crafted features and algorithms~\cite{haeberli1990paint,hertzmann1998painterly,hertzmann2001image,efros2001image}. Gatys et al.~\cite{Gatys_2016_CVPR} introduced Neural Style Transfer and dramatically improved the state-of-the-art by leveraging the features of a pretrained CNN. Neural Style Transfer represents ``style" using the Gram matrix of features extracted from the shallow layers of a CNN and ``content" using feature tensors extracted from the deeper layers. The pixels of a stylized output image are directly optimized to simultaneously match the style representation (of the style image) and content representation (of the content image). Subsequent works improve upon ~\cite{Gatys_2016_CVPR} with different complementary schemes, including spatial constraints~\cite{Selim_2016_ACM}, semantic guidance~\cite{Champandard_2016}, and Markov Random Field priors~\cite{Li_2016_CVPR}. Other work has improved upon ~\cite{Gatys_2016_CVPR} by replacing the objective function, style representation, and/or content representation ~\cite{mechrez2018contextual,gu2018arbitrary,Kolkin_2019_CVPR}.

Optimization-based methods produce high quality stylizations, but they can be computationally expensive as they require backpropagation at every iteration and gradually change the image, usually at a pixel level, until the desired statistics are matched. To overcome this limitation, model-based neural methods were introduced. These methods optimize a generative model offline, and at test time produce a stylized image with a single forward pass. These methods fall into two families with different tradeoffs relative to optimization-based style transfer. Some methods~\cite{sanakoyeu2018styleaware} trade flexibility for speed and quality, quickly producing excellent stylizations but only for a predetermined set of styles. Other methods~\cite{johnson2016perceptual,huang2017arbitrary} trade off quality for speed, allowing for fast transfer of arbitrary styles, but typically produce lower quality outputs than optimization-based style transfer. Each family of method excels in a different regime, and in this work we prioritize flexibility and quality over speed.

Until recently, style transfer methods could not transfer geometric style and were limited to transferring color and texture. In much work outside the domain of style transfer, however, geometric transformation has been applied to images via automatic warping. Early work required predicting a set of global transformation parameters or a dense deformation field. Cole et al.~\cite{Cole_2017_CVPR} enabled fine-grained local warping by proposing a method that takes a sparse set of control keypoints and warps an image with spline interpolation. The introduced warping module is differentiable and thus can be trained as part of an end-to-end system, although~\cite{Cole_2017_CVPR} requires pre-detected landmarks as input for its face synthesis task.

Several recent works have attempted to combine image warping with neural networks to learn both textural and geometric style of human portraits. CariGAN~\cite{Li_2018} translates a photo to a caricature by training a Generative Adversarial Network (GAN) that models geometric transformation with manually annotated facial landmarks and another GAN that translates the usual non-geometric style appearances. Face of Art (FoA)~\cite{Yaniv_2019_ACM} trains a neural network model to automatically detect 68 canonical facial landmarks in artistic portraits and uses them to warp a photo so that the geometry of the face is closer to that of an artistic portrait. WarpGAN~\cite{Shi_2019_CVPR}, on the other hand, adds a warping module to the generator of a GAN and trains it as part of an end-to-end system by optimizing both the locations of keypoints and their displacements, using a dataset that contains a caricature and photo pair for each identity.

The main distinction of our work from these efforts is that ours is not limited to human faces (or any other particular domain) and does not require offline training on a specially prepared dataset. In terms of methodology, FoA separates transferring ``texture" and transferring geometry, while DST transfers them jointly. WarpGAN treats texture and geometry jointly, but it has to learn a warping module with paired examples in the face caricature domain while DST does not. We show in Section~\ref{sec:results} that results of our more general method, even when applied to faces, are competitive with results of these two face-specific methods.

For finding correspondences between images, two recent efforts use CNN-based descriptors to identify matching points between paired images outside the domain of human faces. Fully Convolutional Self-Similarity~\cite{Kim_2019_IEEE} is a descriptor for dense semantic correspondence that uses local self-similarity to match keypoints among different instances within the same object class. Neural Best-Buddies (NBB)~\cite{Aberman_2018_ACM} is a more general method for finding a set of sparse cross-domain correspondences that leverages the hierarchical encoding of features by pre-trained CNNs. We use NBB in our method with post-processing, as described in detail in the next section.

\section{Geometry Transfer via Correspondences} \label{sec:points}

One path for introducing geometric style transfer is establishing spatial associations between the content and style images, and defining a \emph{deformation} that brings the content image into (approximate) alignment with the style image. Assuming they share a domain and have similar geometry (e.g. both are images of front-facing cars), we can aim to find meaningful spatial correspondences to define the deformation. The correspondences would specify displacement ``targets'', derived from the style image, for keypoints in the content image. Thin-plate spline interpolation~\cite{Cole_2017_CVPR} can extend this sparse set of displacements to a full displacement field specifying how to deform every pixel in the output image. 

\begin{figure}
\floatbox[{\capbeside\thisfloatsetup{capbesideposition={left,center},capbesidewidth=5cm}}]{figure}[\FBwidth]
{\caption{Illustration of our method using keypoints taken from FoA (rows 1-3) or generated manually (row 4). Keypoints are overlayed on the content and style images with matching points in the same color. Naive warp indicates output of standard style transfer warped by moving source points to target points. Figure is best viewed zoomed-in on screen.}\label{fig:foapoints}}
{\includegraphics[height=6.5cm]{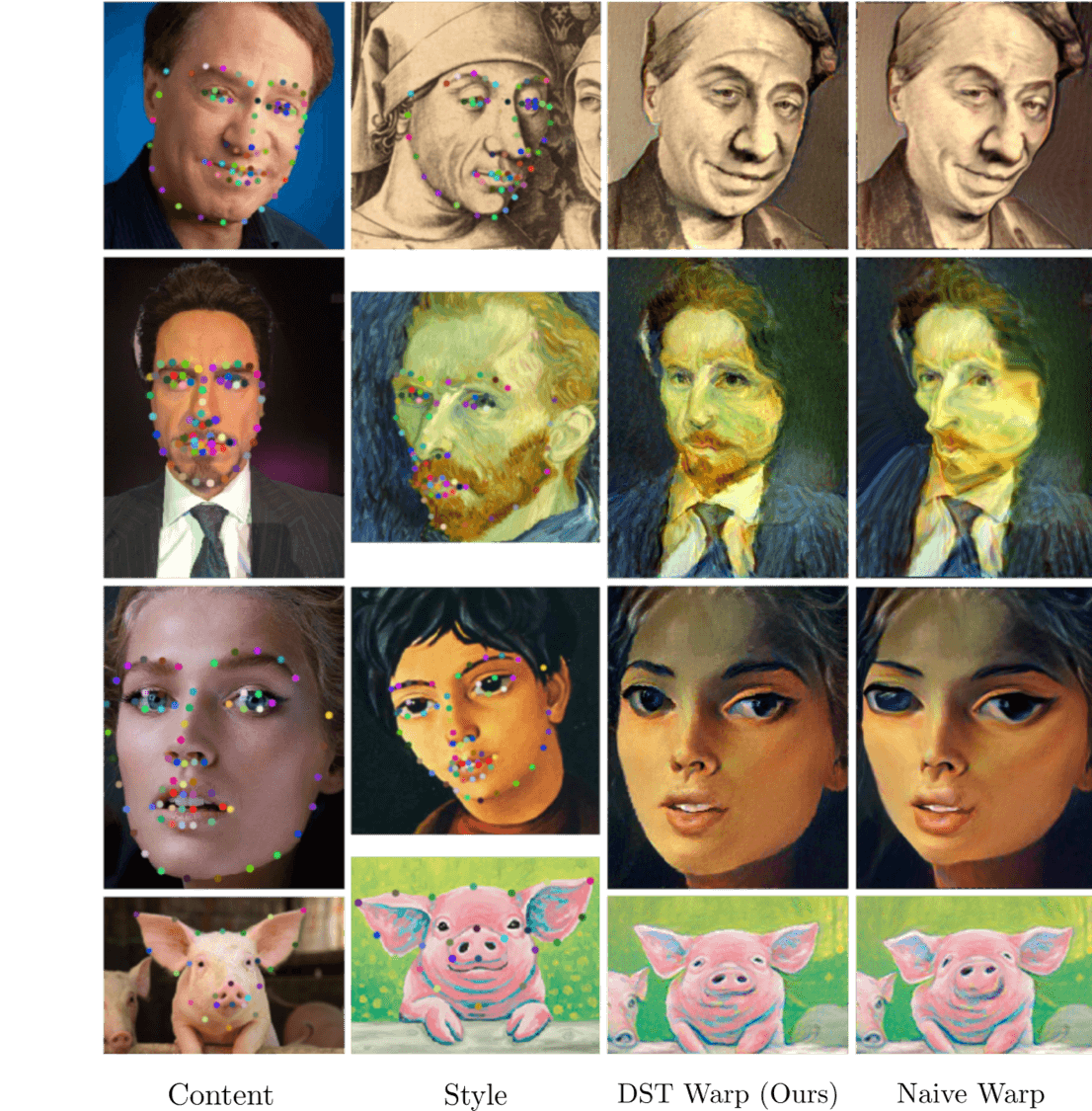}}
\end{figure}

\subsection{Finding and Cleaning Keypoints}
If we fix a domain and assume availability of a training set drawn from the domain, we may be able to learn a domain-specific mechanism for finding salient and meaningful correspondences. This can be done through facial landmark detection~\cite{Yaniv_2019_ACM} or through learning a data-driven detector for relevant points~\cite{Kim_2019_IEEE,Shi_2019_CVPR}. Alternatively, we can expect a user interacting with a style transfer tool to manually select points they consider matching in the two images. If matching points are provided by such approaches, they can be used in DST as we show in Figure~\ref{fig:foapoints}. However, we are interested in a more general scenario, a one-shot, domain-agnostic setting where we may not have access to such points. Hence we use NBB, a generic method for point matching between images. 

NBB finds a sparse set of correspondences between two images that could be from different domains or semantic categories. It utilizes the hierarchy of deep features of a pre-trained CNN, i.e. the characteristic that deeper layers extract high-level, semantically meaningful, and spatially invariant features and shallow layers encode low-level features such as edges and color features. Starting from the deepest layer, NBB searches for pairs of correspondences that are mutual nearest neighbors, filters them based on activation values, and percolates them through the hierarchy to narrow down the search region at each level. At the end of the algorithm, it clusters the set of pixel-level correspondences into $k$ spatial clusters and returns $k$ keypoint pairs.

\begin{figure}[htbp!]
\begin{center}
\begin{tabular}{ccccc}
	\includegraphics[height =0.14\linewidth]{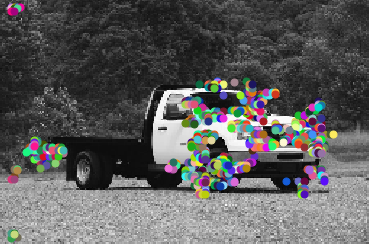} &   
	\includegraphics[height=0.14\linewidth]{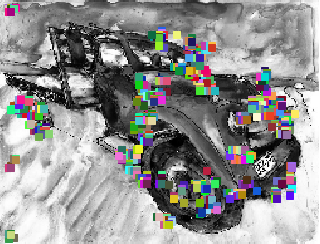} &
	\includegraphics[height=0.14\linewidth]{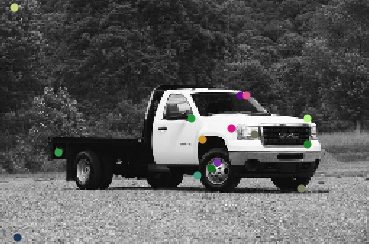} &
	\includegraphics[height=0.14\linewidth]{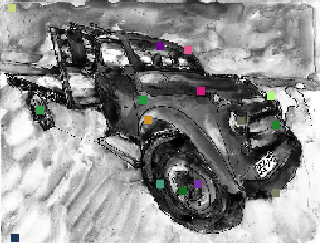} \\
	(a) & (b) & (c) & (d) \\
	\includegraphics[height=0.14\linewidth]{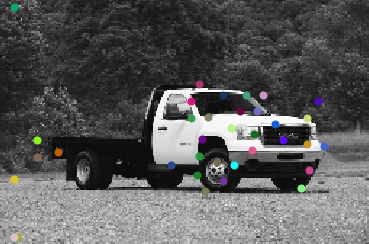} &
	\includegraphics[height=0.14\linewidth]{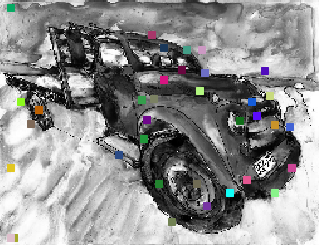} &
	\includegraphics[height=0.14\linewidth]{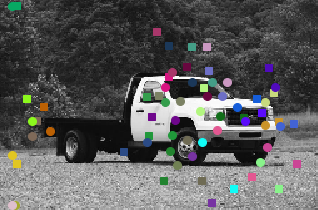} &
     \includegraphics[height=0.14\linewidth]{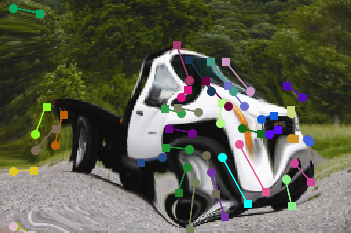} \\
	(e) & (f) & (g) & (h)
\end{tabular}
\end{center}
\vspace{-0.5cm}
\caption{An image can be spatially deformed by moving a set of source points to a set of target points. Matching keypoints are indicated by color. (a) Content image with all correspondences. (b) Style image with all correspondences. (c) Content image with original NBB keypoints. (d) Style image with original NBB keypoints. (e) Content image with our selected keypoints. (f) Style image with our selected keypoints. (g) Content image with keypoints aligned by just matching the centers. (h) Content image warped with keypoints aligned by a similarity transformation. The lines indicate where the circle source points move to (square target points). Figure is best viewed zoomed-in on screen.}
\label{fig:points}
\end{figure}

The keypoint pairs returned by NBB, however, are often too noisy and not sufficiently spread out to use them as they are. To provide better guidance for geometric deformation, we modify NBB to get a cleaner and better spatially-distributed set of pairs. Specifically, we remove the final clustering step and return all pixel-level correspondences, usually on the order of hundreds of correspondence pairs. Then we use a greedy algorithm that selects a keypoint with the highest activation value (calculated by NBB) that is at least $10$ pixels away from any already selected keypoint. We select up to $80$ keypoint pairs and filter out keypoints with activation values smaller than $1$. After the initial selection, we map the keypoints in the style image onto the content image by finding a similarity transformation that minimizes the squared distance between the two point clusters~\cite{Umeyama1991}. We then additionally clean up the selected keypoints by removing keypoints pairs that cross each other, to prevent a discontinuous warp field. (If keypoints are provided by FoA, manual selection, or other non-NBB methods, we skip the cleaning process and only transform the style image keypoints appropriately.) We refer to the keypoints in the content image as the ``source points'' and the corresponding keypoints in the style image mapped onto the content image as the ``target points.'' This process is illustrated in Figure~\ref{fig:points}.

\subsection{Differentiable Image Warping}
We specify an image deformation by a set of source keypoints $P=\{p_1,\ldots,p_k\}$ and the associated 2D displacement vectors $\theta=\{\theta_1,\ldots,\theta_k\}$. $\theta$ specify for each source keypoint $p_i$, the \emph{destination} coordinates $p_i+\theta_i$. Following~\cite{Shi_2019_CVPR} we use thin-plate spline
interpolation~\cite{Cole_2017_CVPR} to produce a dense flow field from the coordinates of an unwarped image $I$ to a warped image $W(I,\theta)$. This is a closed-form procedure which finds parameters $w, v, b$ that minimize $\sum_{i=1}^k ||f_\theta(p_i+\theta_i) - p_i||^2$ subject to a curvature constraint. With these parameters, we have the inverse mapping function
\begin{equation}
f_\theta(q) = \sum_{i=1}^{k} w_i \phi(||q - p_i-\theta_i||) + v^T q + b
\end{equation}
where $q$ denotes the location of a pixel in the warped image and $\phi$ is a kernel function which we choose to be $\phi (r) = r^2 \log(r)$. $f_\theta(q)$ gives the inverse mapping of the pixel $q$ in the original image; i.e. the pixel coordinates in the unwarped image from which we should derive the color of pixel $q$ in the warped image. The color of each pixel can then be generated through bilinear sampling. This entire warping module is differentiable with respect to $\theta$, allowing it to be optimized as part of an end-to-end system.

\section{Spatially Guided Style Transfer}\label{sec:approach}

The input to DST consists of a content image $I_c$, a style image $I_s$, and aligned keypoint pairs $P$ (source) and $P'$ (target). Recall that these points don't have to be infused with explicit domain- or category-specific semantics. DST optimizes the stylization parameters (usually the pixels of the output image) $X$ and the deformation parameters $\theta$. The final output is the warped stylized image $W(X, \theta)$.

\subsection{Content and Style Loss Terms} \label{sec:contentstyle}

DST can be used with any one-shot, optimization-based style transfer method with a content loss and a style loss. In this work, we demonstrate our framework with two such methods: Gatys et al.~\cite{Gatys_2016_CVPR} and Kolkin et al.~\cite{Kolkin_2019_CVPR}, which we will refer to as Gatys and STROTSS, respectively. Each method defines a content loss $\lcon(I_c,X)$ and a style loss $\lsty(I_s,X)$. These aim to capture the visual content of $I_c$ and the visual style of $I_s$ in the output $X$. Below we briefly summarize the content/style loss of these methods. For more details, we direct the reader to~\cite{Gatys_2016_CVPR,Kolkin_2019_CVPR}.

Gatys represents ``style" in terms of the Gram matrix of features extracted from multiple layers of a CNN and ``content" as the feature tensors extracted from another set of layers. It defines $\lcon$ as the squared-error between the feature representation of the content image and that of the output image. Similarly, it defines $\lsty$ as the weighted sum of the squared-error between the Gram matrix of the style image and that of the output image.

STROTSS, inspired by the concept of self-similarity, defines $\lcon$ as the absolute error between the normalized pairwise cosine distance between feature vectors extracted from the content image and those of the output image. STROTSS's $\lsty$ is composed of three terms: the Relaxed Earth Movers Distance (EMD), the moment matching term, and the color matching term. Relaxed EMD helps transfer the structural forms of the style image to the output image. The moment matching term, which aims to match the mean and covariance of the feature vectors in the two images, combats over-or under-saturation. The color matching term, defined as the Relaxed EMD between pixel colors, encourages the output image and the style image to have a similar palette.

When using DST with a base style transfer method, we do not change anything about $\lcon$. The style loss of DST is composed of two terms
\begin{equation}
\lsty(I_s, X) + \lsty(I_s, W(X, \theta)).\label{eq:lsty}
\end{equation}
The first loss term is between the style image $I_s$ and the \emph{unwarped} stylized image $X$. The second term is between $I_s$ and the spatially deformed stylized image $W(X, \theta)$, with $\theta$ defining the deformation as per Section~\ref{sec:points}. Minimizing Eq.~\eqref{eq:lsty} is aimed at finding a good stylization both with and without spatial deformation. This way we force the stylization parameters $X$ and the spatial deformation parameters $\theta$ to work together to produce a harmoniously stylized and spatially deformed final output $W(X, \theta)$.

\subsection{Deformation Loss Term}
Given a set of $k$ source points $P$ and matching target points $P'$, we define the deformation loss as
\begin{equation}\label{eq:lmatch}
  \lmatch(P,P',\theta) = \frac{1}{k}\sum_{i=1}^k \| p'_i - (p_i+\theta_i)\|_2,
\end{equation}
where $p_i$ and $p_i'$ are the $i$-th source and target point coordinates. Minimizing Eq.~\eqref{eq:lmatch} with respect to $\theta$ seeks a set of displacements that move the source points to the target points. This term encourages the geometric shape of the stylized image to become closer to that of the style.

Aggressively minimizing the deformation loss may lead to significant artifacts, due to keypoint match errors or incompatibility of the content with the style geometry. To avoid these artifacts, we add a regularization term encouraging smooth deformations. Specifically, we use the (anisotropic) total variation norm of the 2D warp field $f$ normalized by its size
\begin{equation}
  R_\text{TV}(f) = \frac{1}{\text{W}\times \text{H}} \sum_{i=1}^{\text{W}} \sum_{j=1}^{\text{H}} \|f_{i+1, j} - f_{i, j}\|_1 +  \|f_{i, j+1} - f_{i, j}\|_1.
\end{equation}
This regularization term smooths the warp field by encouraging nearby pixels to move in a similar direction. 

\subsection{Joint Optimization}

\begin{figure}[htbp!]
\begin{center}
\includegraphics[width=0.85\linewidth]{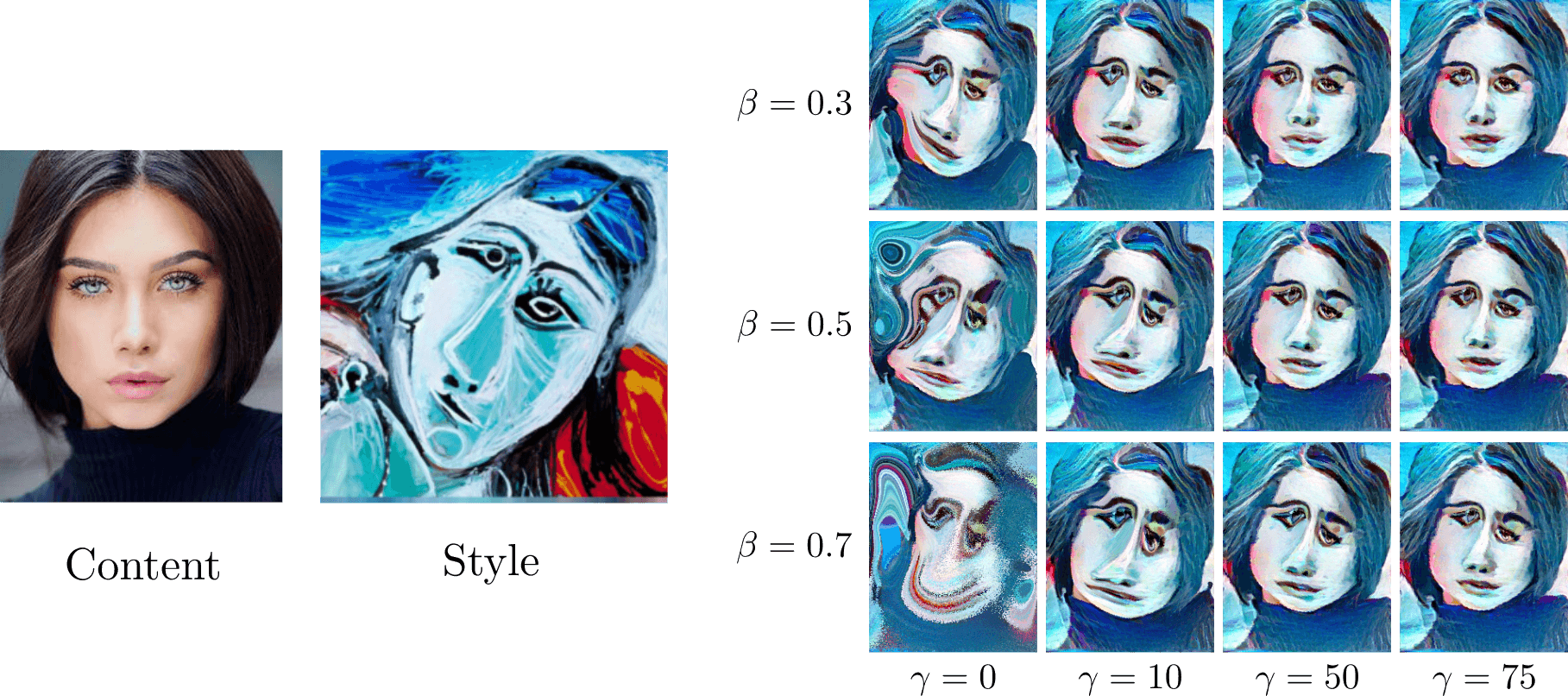}
\caption{DST outputs with varying $\beta$ and $\gamma$ using STROTSS as the base method. Image in the upper right corner (low $\beta$, high $\gamma$) has the least deformation, and the image in the bottom left corner (high $\beta$, low $\gamma$) has the most deformation.} \label{fig:gammabeta}
\end{center}
\end{figure}

Putting everything together, the objective function of DST is
\begin{align}
  L(X, \theta, I_c, I_s, P, P') \label{eq:obj}
  &=\,\alpha \lcon(I_c,X) \\
  &\phantom{=}\,+\, \lsty(I_s,X)\,+\,\lsty(I_s,W(X,\theta)) \nonumber \\
  &\phantom{=}\,+\, \beta\lmatch(P, P', \theta) \nonumber \\
  &\phantom{=}\,+\, \gamma R_\text{TV}(f_\theta), \nonumber
\end{align} 
where $X$ is the stylized image and $\theta$ parameterizes the spatial deformation. Hyperparameters $\alpha$ and $\beta$ control the relative importance of content preservation and spatial deformation to stylization. Hyperparameter $\gamma$ controls the amount of regularization on the spatial deformation. The effect of varying $\alpha$ is analyzed in~\cite{Gatys_2016_CVPR,Kolkin_2019_CVPR}. The effect of changing $\beta$ and $\gamma$ is illustrated in Figure~\ref{fig:gammabeta}. We use standard iterative techniques such as stochastic gradient descent or L-BFGS to minimize Eq.~\eqref{eq:obj} with respect to $X$ and $\theta$. Implementation details can be found in the supplementary material and the published code. \footnote{\texttt{https://github.com/sunniesuhyoung/DST}}

\section{Results}\label{sec:results}

\begin{figure*}
\begin{center}
	\includegraphics[width=0.92\textwidth]{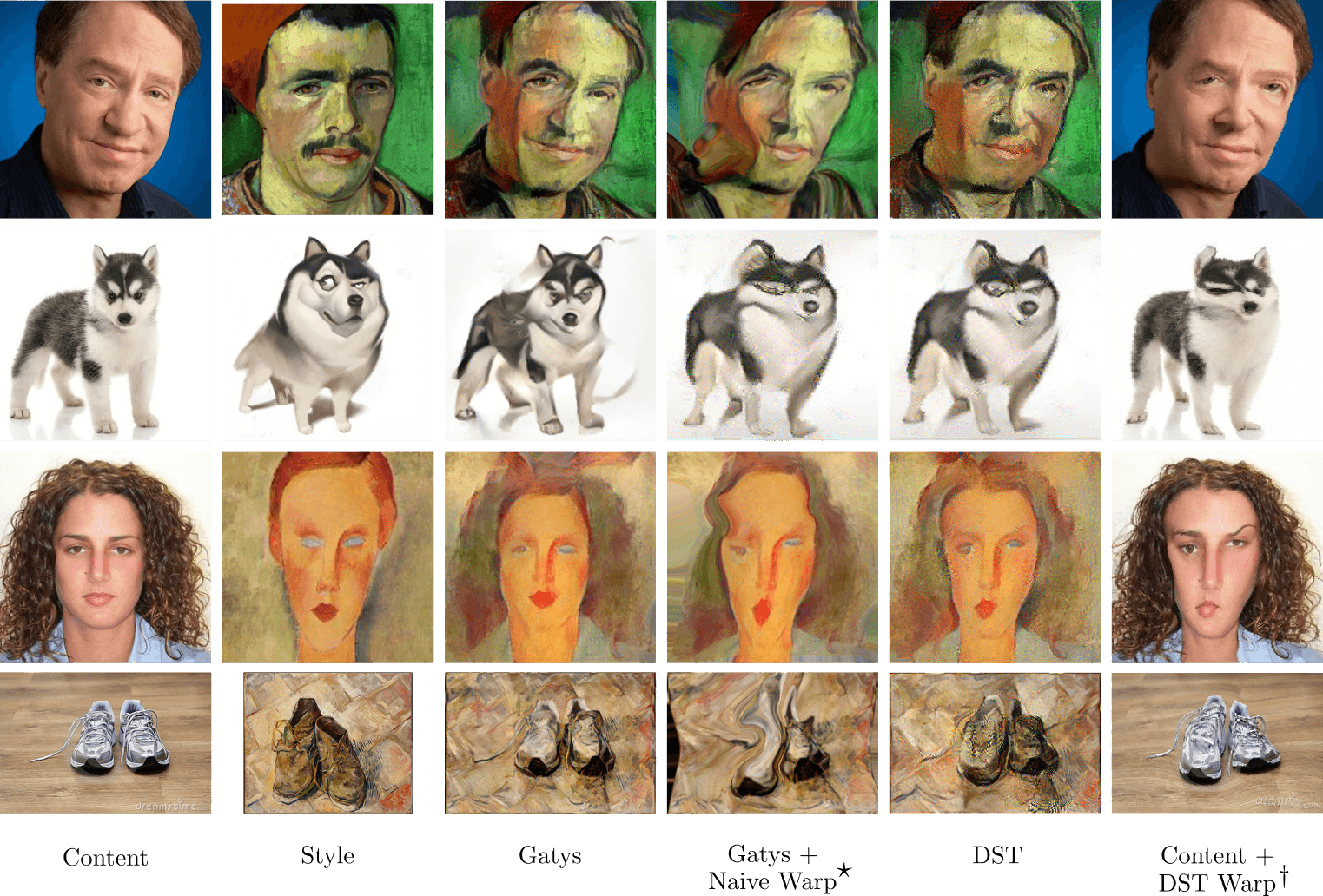}
\end{center}
\caption{DST results with Gatys.$~^{\star}$Naively warped by moving source points to target points.$~^{\dag}$Warp learned by DST applied to the content image.}
\label{fig:gatysresults}
\end{figure*}

\begin{figure*}
\begin{center}
  \includegraphics[width=0.96\textwidth]{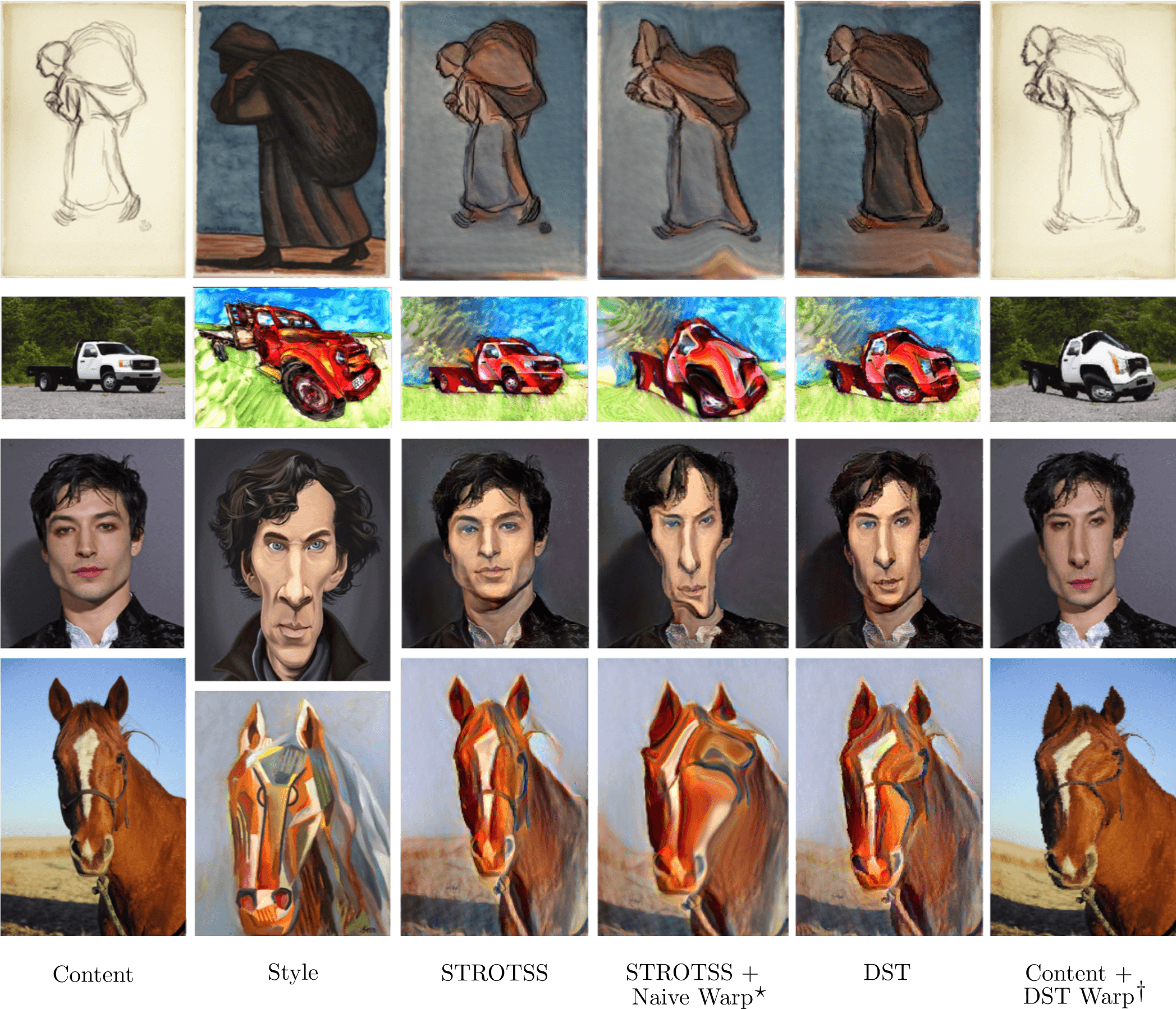}
\end{center}
\caption{DST results with STROTSS.$~^{\star}$Naively warped by moving source points to target points.$~^{\dag}$Warp learned by DST applied to the content image.}
\label{fig:strotssresults}
\end{figure*}

We observe that DST often captures the geometric style of the target style image. One visually striking effect is that the resulting images no longer look like ``filtered'' versions of the original content, as they often do with standard style transfer methods. We show results of DST with Gatys and STROTSS in Figures~\ref{fig:gatysresults} and ~\ref{fig:strotssresults}. For a pair of content and style images, we show the output of DST and the output of Gatys/STROTSS that doesn't have the spatial deformation capability. To highlight the effect of the DST-learned deformation, we also provide the content image warped by DST and the Gatys/STROTSS output naively warped with the selected keypoints without any optimization of the deformation. While naive warping produces undesirable artifacts, DST finds a warp that harmoniously improves stylization while preserving content. 

As a simple quantitative evaluation, we calculated the (STROTSS) style loss on 185 pairs of DST and STROTSS outputs. Surprisingly, we found that on average this loss was ~7\% higher for DST outputs than STROTSS ones, even for examples we show in Figure~\ref{fig:strotssresults}. While the loss difference is small, this is a mismatch with the human judgment of stylization quality shown in Section~\ref{sec:mturk}.

\subsection{Comparison with FoA and WarpGAN}\label{sec:facecompare}

\begin{figure}[htbp!]
\begin{center}
	\begin{tabular}{cc}
	\multicolumn{2}{c}{\begin{tabular}{cccc|cccc}
	\includegraphics[height=0.115\linewidth]{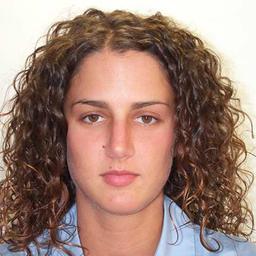} &
	\includegraphics[height=0.115\linewidth]{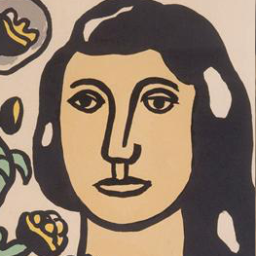} &
	\includegraphics[height=0.115\linewidth]{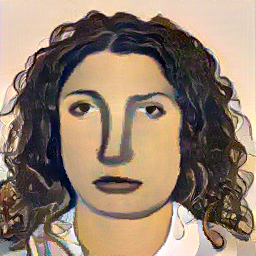} &
	\includegraphics[height=0.115\linewidth]{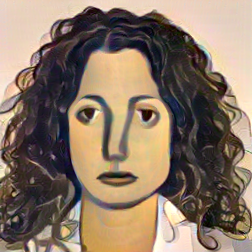} &
	\includegraphics[height=0.115\linewidth]{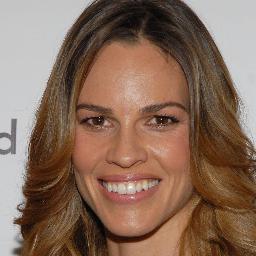} &
	\includegraphics[height=0.115\linewidth]{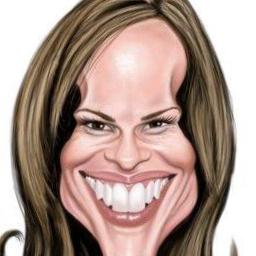} &
	\includegraphics[height=0.115\linewidth]{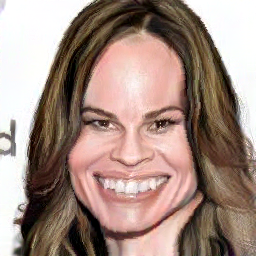} &
	\includegraphics[height=0.115\linewidth]{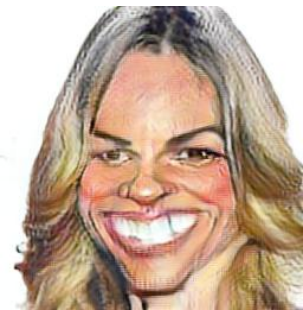} \\
	\includegraphics[height=0.115\linewidth]{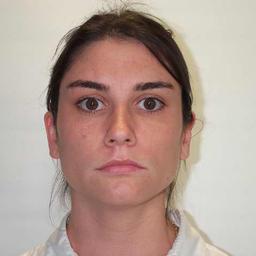} &
	\includegraphics[height=0.115\linewidth]{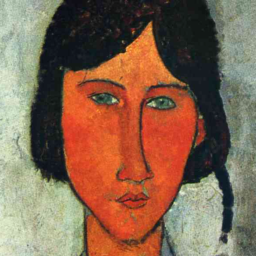} &
	\includegraphics[height=0.115\linewidth]{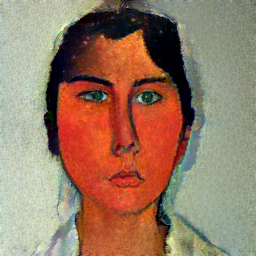} &
	\includegraphics[height=0.115\linewidth]{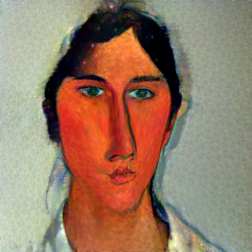} &
	\includegraphics[height=0.115\linewidth]{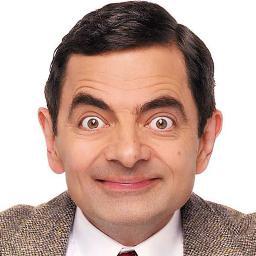}  &
	\includegraphics[height=0.115\linewidth]{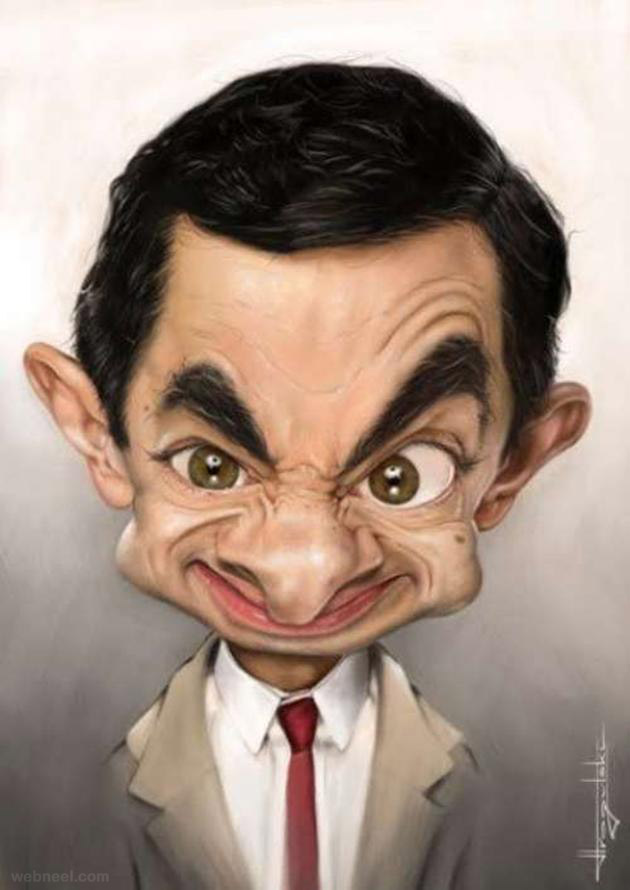} &
	\includegraphics[height=0.115\linewidth]{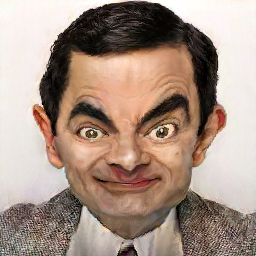} &
	\includegraphics[height=0.115\linewidth]{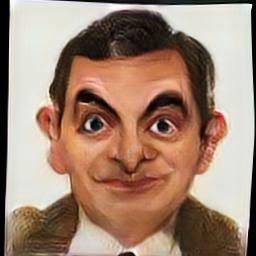}\\
	\includegraphics[height=0.115\linewidth]{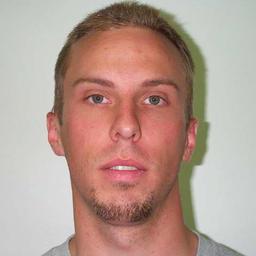} &
	\includegraphics[height=0.115\linewidth]{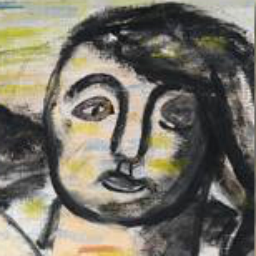} &
	\includegraphics[height=0.115\linewidth]{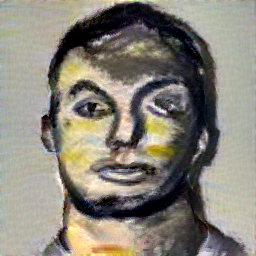} &
	\includegraphics[height=0.115\linewidth]{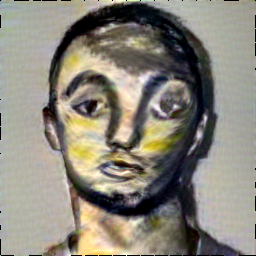}&
  	\includegraphics[height=0.115\linewidth]{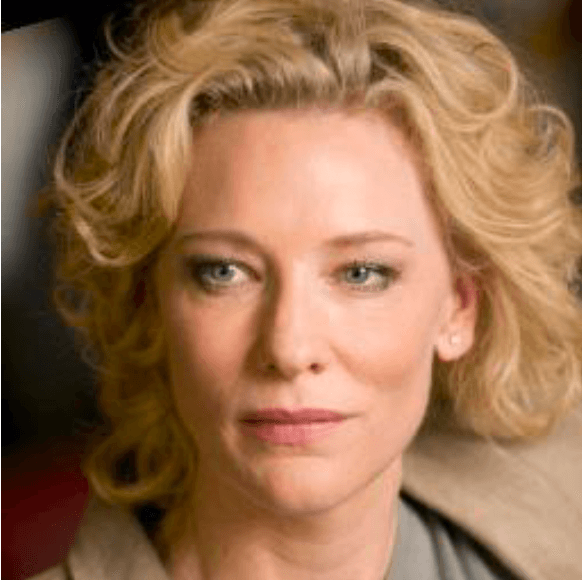}  &
	\includegraphics[height=0.115\linewidth]{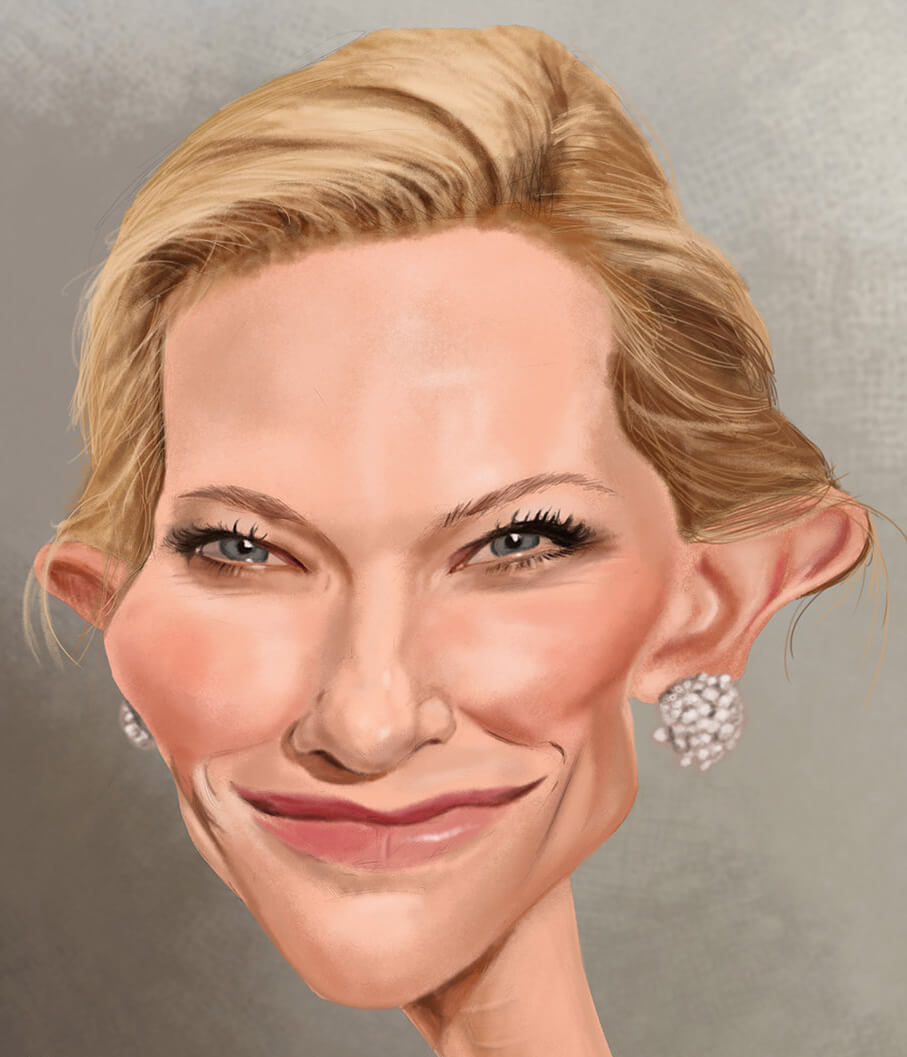} &
	\includegraphics[height=0.115\linewidth]{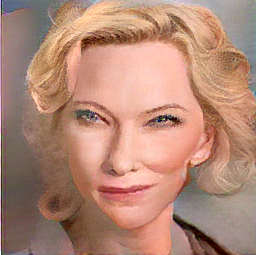} &
	\includegraphics[height=0.115\linewidth]{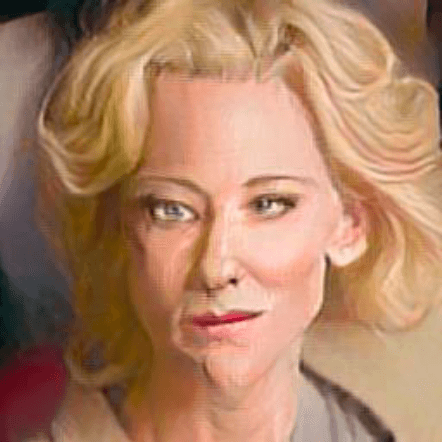}\\
	Content & Style & DST & FoA  & Content & Style$^\dagger$ & DST & WarpGAN\\
	\end{tabular}}\\
\hspace{0.23\linewidth}(a) & \hspace{0.21\linewidth}(b)
\end{tabular}
\caption{Comparison of DST with (a) FoA and (b) WarpGAN. $\dagger$: Note that WarpGAN's does not use a specific style image, so this style image is only used by DST; see text for details.}\label{fig:foacompare}
\end{center}
\end{figure}

While, so far as we are aware, DST is the first work to allow open-domain geometry-aware style transfer, other work has tackled this problem in the domain of human faces. To compare performance, we show results of DST and results of FoA~\cite{Yaniv_2019_ACM} and WarpGAN~\cite{Shi_2019_CVPR} on the same content-style pairs in Figure~\ref{fig:foacompare}. Note that both of these methods require training a model on a dataset of stylized portraits or caricatures, while DST operates with access to only a single content and single style image.

DST jointly optimizes the geometric and non-geometric stylization parameters, while FoA transfers geometric style by warping the facial landmarks in the content image to a specific artist's (e.g. Modigliani) canonical facial landmark pattern (with small variations added) learned by training a model on a dataset of stylized portraits. FoA then separately transfers textural style with a standard style transfer method (e.g. Gatys, STROTSS). When we compare DST and FoA in Figure~\ref{fig:foacompare}, we demonstrate ``one-shot FoA" since the style images used to produce the outputs in~\cite{Yaniv_2019_ACM} are unavailable. That is, we assume that we have access to one content image, one style image, and the trained FoA landmark detector. Using the detector, we find 68 facial landmarks in the content and style images and transform the style image landmarks, as described in Section~\ref{sec:points}, to get the target points. Then we follow FoA's two-step style transfer and transfer the textural style by STROTSS and transfer the geometric style by warping the output image by moving the source points to the target points. 

The biggest difference between WarpGAN and DST is that DST is a one-shot style transfer method that works with a single content image and a single style image. WarpGAN, on the other hand, is trained on a dataset of paired pictures and caricatures of the same identities, and generates a caricature for an input content image from its learned deformation model. To compare the performance of WarpGAN and DST, we used content/style image pairs from~\cite{Shi_2019_CVPR} and ran DST. In Figure~\ref{fig:foacompare}, we show the outputs of DST and the outputs of WarpGAN taken from~\cite{Shi_2019_CVPR}. Despite the lack of a \emph{learning} component, DST results are competitive and sometimes more aesthetically pleasing than results of FoA and WarpGAN.

\subsection{Human Evaluation}\label{sec:mturk}
Quantitatively evaluating and comparing style transfer is challenging, in part because of the subjective nature of aesthetic properties defining style and visual quality, and in part due to the inherent tradeoff between content preservation and stylization~\cite{yeh2018quantitative,Kolkin_2019_CVPR}. Following the intuition developed in these papers, we conducted a human evaluation study using Amazon Mechanical Turk on a set of 75 diverse style/content pairs. The goal was to study the effect of DST on the stylization/content preservation tradeoff, in comparison to the base style transfer methods. The evaluation was conducted separately for STROTSS and Gatys-based methods. We considered three DST deformation regimes: low ($\beta=0.3$,$\gamma=75$), medium ($\beta=0.5$,$\gamma=50$), and high ($\beta=0.7$,$\gamma=10$) for STROTSS;  low ($\beta=3$,$\gamma=750$), medium ($\beta=7$,$\gamma=100$), and high ($\beta=15$,$\gamma=100$) for Gatys. So for each base method, we compare four stylized output images. The effect of varying $\beta$ and $\gamma$ is illustrated in Figure~\ref{fig:gammabeta}.

To measure content preservation, we asked MTurk users the question: ``Does image A represent the same scene as image B'', where A referred to the content image and B to the output of style transfer. The users were forced to choose one of four answers: ``Yes'', ``Yes, with minor errors'', ``Yes, with major errors'' and ``No''. Converting these answers to numerical scores (1 for ``No'', 4 for ``Yes'') and averaging across content/style pairs and users, we get a \emph{content score} between 1 and 4 for each of the four methods.

To evaluate the effect of the proposed deformable framework, we presented the users with a pair of outputs, one from the base method (Gatys or STROTSS) and the other from DST, along with the style image. The order of the first two is randomized. We asked the users to choose which of the two output images better matches the style. The fraction of time a method is preferred in all comparisons (across methods compared to, users, content/style pairs) gives a \emph{style score} between 0 and 1. 0.7 means that the method ``wins'' 70\% of all comparisons it was a part of. The evaluation interfaces are provided in the supplementary material.

In total, there were 600 unique content comparisons: 4 questions$\times$75 images for Gatys and an equal number for STROTSS. 123 users participated in the evaluation, and each comparison was evaluated by 9.55 users on average. The standard deviation of the content choice agreement was 0.79 (over a range of 1 to 4). For stylization, there were 450 unique comparisons in total: 3 comparisons between the base method and each of the 3 DST deformation regimes$\times$75 images for Gatys and likewise for STROTSS. 103 users participated in the stylization evaluation, and each comparison was evaluated by 8.76 users on average. For each comparison, 6.47 users agreed in their choice on average.

\begin{figure}
\floatbox[{\capbeside\thisfloatsetup{capbesideposition={right,center},capbesidewidth=5cm}}]{figure}[\FBwidth]
{\caption{Human evaluation results, comparing DST in different deformation regimes with STROTSS (green) and Gatys (blue). DST provides a much higher perceived degree of style capture without a significant sacrifice in content preservation.}\label{fig:mturk} }
{\includegraphics[height=6.5cm]{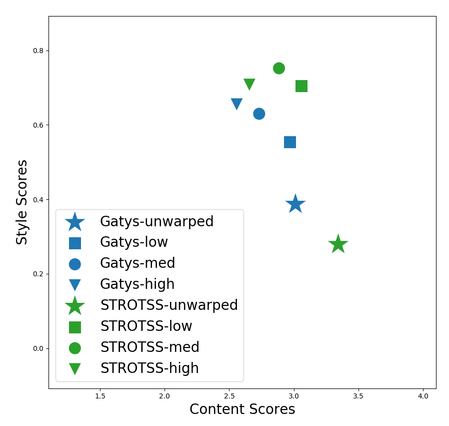}}
\end{figure}

Results of this human evaluation are shown in Figure~\ref{fig:mturk}. Across the deformation regimes (low, medium, high), for both STROTSS and Gatys, DST significantly increases the perceived stylization quality, while only minimally reducing the perceived content preservation. Note that some reduction to the content score can be expected since we intentionally alter the content more by deforming it, but our evaluation shows that this drop is small.

\subsection{Limitations}
\begin{figure}
\floatbox[{\capbeside\thisfloatsetup{capbesideposition={right,center},capbesidewidth=5.5cm}}]{figure}[\FBwidth]
{\caption{Examples of DST failures. We observed that stylization failures are often due to correspondence errors or overly complex scene layout.}\label{fig:failures}}
{\includegraphics[height=4.6cm]{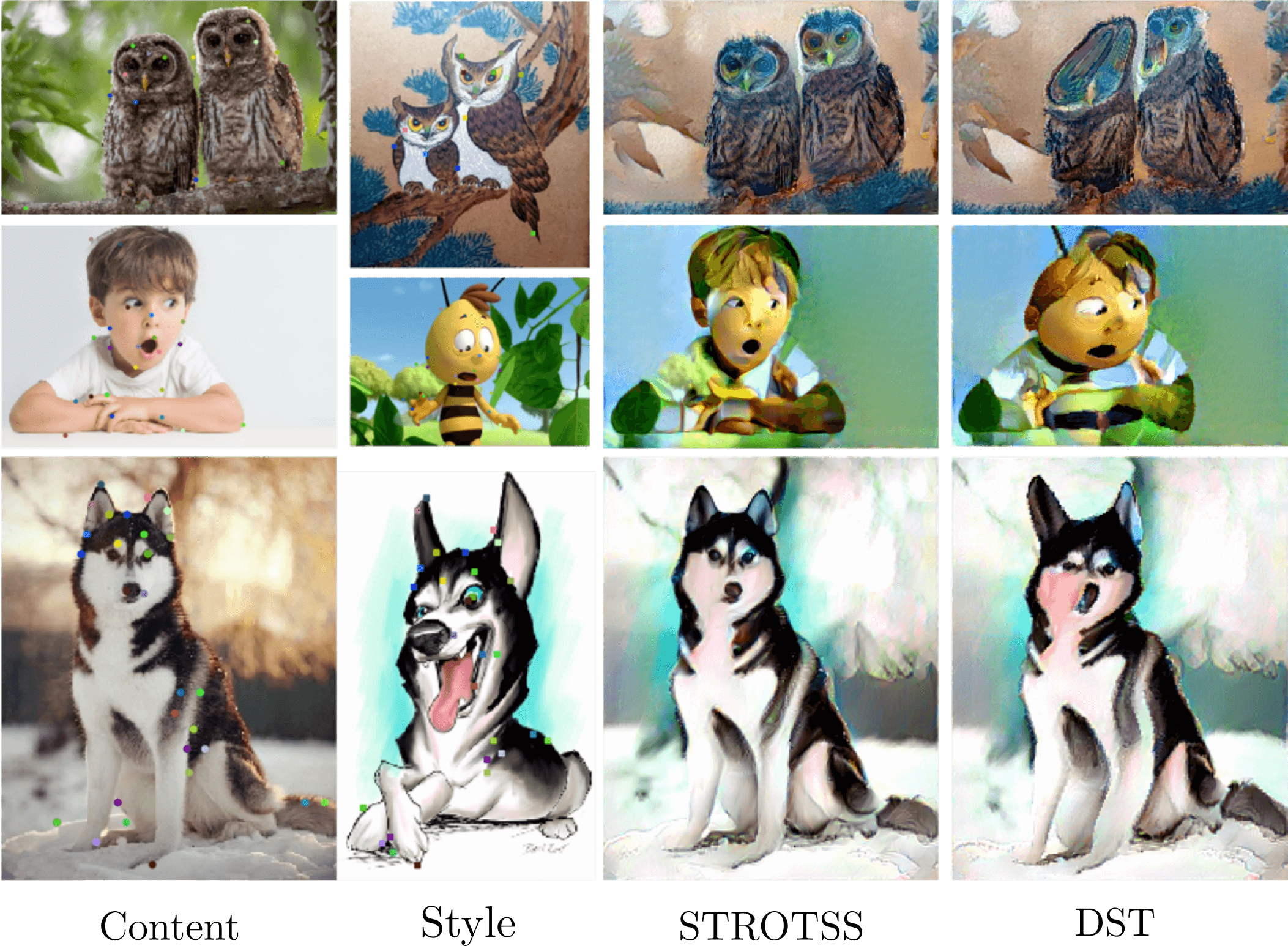}}
\end{figure}

In Figure~\ref{fig:failures}, we show unsuccessful examples of DST where the output image did not deform towards having a similar shape as the style image or deformed only partially. We observed that bad deformations often stem from poorly matching or too sparse set of keypoints. We expect finding better matching keypoints between images and making the method more robust to poor matches will improve results.

\section{Conclusion}\label{sec:conclusions}

Prior work on style transfer has largely ignored geometry and shape, despite the role these play in visual style. We present deformable style transfer (DST), a novel approach that combines the traditional texture and color transfer with spatial deformations. Our method incorporates deformation targets, derived from domain-agnostic point matching between content and style images, into the objective of an optimization-based style transfer framework. This is to our knowledge the first effort to develop a one-shot, domain-agnostic method for capturing and transferring geometric aspects of style.

Still many aspects of geometric style transfer remain unexplored. Narrowing this might involve developing more robust keypoint matching algorithms for highly stylized images. Furthermore, it is by no means certain that modifying geometry using warp fields driven by paired keypoints is the most effective approach to this problem. We hope that future work will continue to explore how to represent geometric style more flexibly, and more accurately encode the artistic extraction and abstraction of shape and form.

\section*{Acknowledgement}

This work is supported by the DARPA GARD award HR00112020003 and the University of Chicago CDAC Discovery Grant. We would also like to thank Haochen Wang, Davis Gilton, Steven Basart, and members of the Perception and Learning Systems (PALS) at TTIC for their helpful comments. 

\clearpage
%
%
\bibliographystyle{splncs04}
\bibliography{paper}

\newpage

\section*{Supplementary Material}

\renewcommand{\figurename}{Supplementary Fig.}
\setcounter{figure}{0}

\subsection*{Implementation Details}

To produce DST results with \textbf{Gatys}, we initialized the output image as the content image resized to have a long side of 256 pixels. During the optimization, we updated the parameters 100 times with a learning rate of 1 with L-BFGS, following the authors' implementation settings. Since the content and style loss terms and the deformation loss term have vastly different magnitudes, we scaled down the content loss and the style loss by $1/50000$ and $1/100000$, respectively. We also scaled down the deformation parameter gradients by $1/1000000$ to update them at a much smaller rate. 

To produce DST results with \textbf{STROTSS}, we followed the authors' implementation settings and optimized the output image at multiple scales, starting from a low-resolution image scaled to have a long side of 64 pixels. The output image was initialized as the bottom level of a Laplacian pyramid of the content image added to the mean color of the style image. During the optimization, instead of optimizing the pixels directly, we optimized the Laplacian pyramid of the image for faster convergence, in addition to the deformation parameters $\theta$. At each scale, we used the output of the previous scale as the initial image, used bilinear upsampling to increase the resolution, and halved the content weight $\alpha$ that controls the relative importance of content preservation to stylization. We produced the stylized images in the main paper at three scales, starting from content weight $\alpha = 32$. At each scale, we made 350 updates of the parameters using stochastic gradient descent with a learning rate of 0.2. More details can be found in our published code: \texttt{https://github.com/sunniesuhyoung/DST}.

\subsection*{Human Evaluation}

As described in Section~\ref{sec:mturk} of the main paper, we conducted a human evaluation study using Amazon Mechanical Turk on a set of 75 diverse style/content pairs. We show the evaluation interfaces in Supplementary Figures~\ref{fig:contenteval} and \ref{fig:styleeval}.

\begin{figure*}
\begin{center}
	\includegraphics[width=0.99\linewidth]{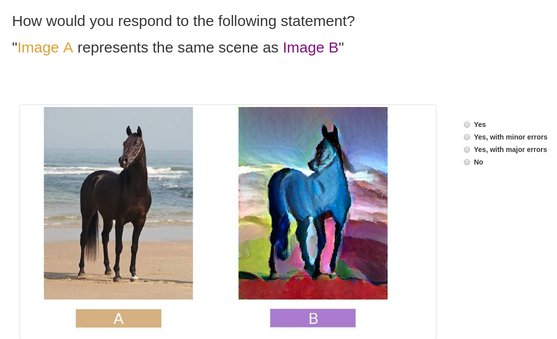}
\end{center}
\caption{Evaluation interface for measuring content preservation.}
\label{fig:contenteval}
\end{figure*}

\begin{figure*}
\begin{center}
	\includegraphics[width=0.99\linewidth]{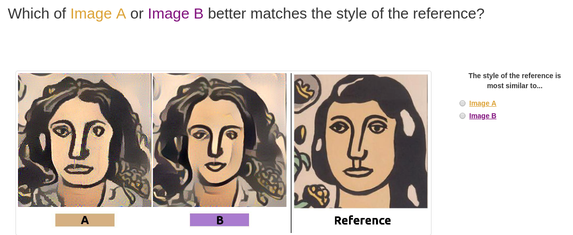}
\end{center}
\caption{Evaluation interface for measuring stylization.}
\label{fig:styleeval}
\end{figure*}

\subsection*{Image Credits}
\subsubsection{Figure 1}
\begin{enumerate}
\item Schoeller, Martin: `Meryl Streep'. (2018), http://www.artnet.com/artists/martin-schoeller/meryl-streep-from-the-series-portraits-a-iuZXKIjNqXY2M9v6ih-p1A2
\item Matisse, Henri: `Girl with a Black Cat'. (1910), https://www.wikiart.org/en/henri-matisse/girl-with-a-black-cat
\item Spellman, Jim: `Jennifer Lawrence Photo'. (2015), https://www.gettyimages.com/detail/news-photo/actress-jennifer-lawrence-attends-the-the-hunger-games-news-photo/497773150
\item Snow, Rob: `Gillian Anderson Caricature'. https://www.juniqe.com/gillian-anderson-premium-poster-portrait-990829.html
\item Bernard, Jean: `Standing Bull Painting'. (c. 1775-1883), \newline https://www.rawpixel.com/image/481523/free-illustration-image-bull-painting-animal
\item Picasso, Pablo: `Bull'. (1945), \newline https://www.artyfactory.com/art\_appreciation/animals\_in\_art/pablo\_picasso.htm
\item Turk, Greg. Levoy, Marc: `Stanford Bunny Render'. (1994), \newline https://en.wikipedia.org/wiki/Stanford\_bunny
\item Walter, Sonia: `Bunny Painting'. https://www.etsy.com/sg-en/listing/533084025/bunny-painting-rabbit-painting-childrens
\item Original artist unknown: `Photo of Man with Shaven Head'. http://www.faculty.idc.ac.il/arik/site/foa/face-of-art.asp
\item Leger, Fernand: `La Botte de Navets'. (1951), \newline https://www.flickr.com/photos/32215553@N02/45868933822
\end{enumerate}

\subsubsection{Figure 2}
\begin{enumerate}
\item Original artist unknown: `Ray Kurzweil Photo', https://www.raadfest.com/ray-kurzweil
\item Meckenem, Israhel van: `Double Portrait of Israhel van Meckenem and His Wife Ida'. (1490), https://www.nga.gov/collection/art-object-page.3311.html
\item Original artist unknown: `Photo of Man in Suit', Unknown Source
\item Van Gogh, Vincent: `Self-Portrait'. (1889), https://www.nga.gov/collection/art-object-page.106382.html
\item Original artist unknown: `Closeup Photo of Blond Woman', https://cgm.technion.ac.il/Computer-Graphics-Multimedia/Software/Contextual/resources/comparison/index.html
\item Kisling, Moise: `Unidentified Portrait', https://www.wikiart.org/en/moise-kisling/not-identified-9/
\item Original artist unknown: `Pig Photo', \newline https://foodsafetynewsfullservice.marlersites.com/files/2016/04/pig-big-ears-406.jpg
\item Original artist unknown: `Poppy the Pig', https://social-artworking.com/poppy-the-pig
\end{enumerate}

\subsubsection{Figure 3}
\begin{enumerate}
\item Original artist unknown: `Truck Photo'. \newline https://marvel-b1-cdn.bc0a.com/f00000000067090/www.constructionequipment.com\newline/sites/ce/files/styles/large/public/X11GM\_SH051\%5B1\%5D.jpg
\item Galaich, George: `Distorted Truck'. https://pixels.com/featured/distorted-truck-george-galaich.html
\end{enumerate}

\subsubsection{Figure 4}
\begin{enumerate}
\item Original artist unknown: `Closeup Photo of Woman with Dark Hair'. \newline https://cgm.technion.ac.il/Computer-Graphics-Multimedia/Software/Contextual/\newline resources/comparison/index.html
\item Picasso, Pablo: `Nude Woman With Necklace'. (1968), https://www.tate.org.uk/art/artworks/\newline picasso-nude-woman-with-necklace-t03670
\end{enumerate}

\subsubsection{Figure 5}
\begin{enumerate}
\item Original artist unknown: `Ray Kurzweil Photo', https://www.raadfest.com/ray-kurzweil
\item Van Gogh, Vincent: `The Zouave'. (June 1888), \newline https://www.vangoghmuseum.nl/en/collection/s0067V1962
\item Original artist unknown: `Husky Puppy Photo', Unknown source
\item Hannson, Stefan: `Husky Caricature'. (2017), https://www.artstation.com/artwork/ry8dm
\item Original artist unknown: `Photo of Woman with Wavy Hair'. \newline http://www.faculty.idc.ac.il/arik/site/foa/face-of-art.asp
\item Modigliani, Amedeo: `Girl with a Polka-Dot Blouse'. (1919), \newline https://collection.barnesfoundation.org/objects/5538/Girl-with-a-Polka-Dot-Blouse-(Jeune-fille-au-corsage-a-pois)/
\item Original artist unknown: `Running Shoes Photo'. http://www.stockphotos.ro/running-shoes-gym-floor-after-workout-image12693027.html
\item Van Gogh, Vincent: `Shoes'. (1888), https://www.metmuseum.org/art/collection/search/436533
\end{enumerate}

\subsubsection{Figure 6}
\begin{enumerate}
\item Original artist unknown: `Peasant Woman Carrying A Pack On Her Back'. https://www.davidsongalleries.com/sections/modern/french-prints/additional-available-works/peasant-woman-carrying-a-pack-on-her-back/
\item Rivera, Diego: `Peasant Woman'. (1946), https://i2.wp.com/blog.dma.org/wp-content/uploads/2017/02/1985\_16\_w.jpg
\item Original artist unknown: `Truck Photo'. \newline https://marvel-b1-cdn.bc0a.com/f00000000067090/www.constructionequipment.com\newline/sites/ce/files/styles/large/public/X11GM\_SH051\%5B1\%5D.jpg
\item Galaich, George: `Distorted Truck'. https://pixels.com/featured/distorted-truck-george-galaich.html
\item Storm, Derek: `Ezra Miller Photo'. (2016), https://www.alamy.com/stock-photo-new-york-ny-usa-10th-nov-2016-ezra-miller-at-arrivals-for-fantastic-125688242.html
\item Snow, Rob: `Benedict Cumberbatch Caricature'.\newline https://cargocollective.com/robsnow\_creative/Caricatures/filter/human/Celebrity-Sunday-Benedict-Cumberbatch
\item Original artist unknown: `Horse Photo'. \newline https://twitter.com/Bbelisle7/status/1215411108921724929/photo/1
\item Sierrra, Dania: `Grey Mane'. https://www.saatchiart.com/art/Painting-Grey-Mane/561637/6472363/view
\end{enumerate}

\subsubsection{Figure 7}
\begin{enumerate}
\item Original artist unknown: `Photo of Woman with Wavy Hair'. \newline http://www.faculty.idc.ac.il/arik/site/foa/face-of-art.asp
\item Leger, Fernand: `La Femme et la fleur'. (1954), https://www.artsy.net/artwork/fernand-leger-la-femme-et-la-fleur
\item Original artist unknown: `Photo of Woman with Pulled Back Hair'. \newline http://www.faculty.idc.ac.il/arik/site/foa/face-of-art.asp
\item Modigliani, Amedeo: `Gypsy Woman with Baby'. (1919), https://www.nga.gov/collection/art-object-page.46649.html
\item Original artist unknown: `Photo of Man with Short Hair'. \newline http://www.faculty.idc.ac.il/arik/site/foa/face-of-art.asp
\item Leger, Fernand: `Male Portrait, http://www.faculty.idc.ac.il/arik/site/foa/face-of-art.asp
\item Original artist unknown: `Photo of Hilary Swank', https://cs.nju.edu.cn/rl/WebCaricature.htm
\item Original artist unknown: `Caricature of Hilary Swank', https://cs.nju.edu.cn/rl/WebCaricature.htm
\item Original artist unknown: `Photo of Rowan Atkinson', https://cs.nju.edu.cn/rl/WebCaricature.htm
\item Original artist unknown: `Photo of Rowan Atkinson', https://cs.nju.edu.cn/rl/WebCaricature.htm
\item Original artist unknown: `Photo of Cate Blanchett', https://cs.nju.edu.cn/rl/WebCaricature.htm
\item Original artist unknown: `Caricature of Cate Blanchett', https://cs.nju.edu.cn/rl/WebCaricature.htm
\end{enumerate}

\subsubsection{Figure 9}
\begin{enumerate}
\item Original artist unknown: `Photo of Barred Owls'. \newline https://1.bp.blogspot.com/-1weEwEzz2wk/Xb64zZC-2RI/\newline AAAAAAADI-M/pTeSYN7v\_KAo2YK3FsPUt6SVe76A5pYaQCLcBGAsYHQ/s640/88l.jpg
\item Yun, Sungsil: `Korean Traditional Watercolor Owl'. https://www.saatchiart.com/art/Painting-Korean-traditional-watercolor-owl/1143968/4610489/view
\item Original artist unknown: `Surprised Boy Photo'.\newline http://weknowyourdreams.com/single/surprise/surprise-01
\item Original artist unknown: `Maya the Bee Render'. https://www.independent.co.uk/arts-entertainment/tv/news/netflix-penis-cartoon-penis-episode-maya-the-bee-a7958551.html
\item Original artist unknown: `Husky Photo'. Unknown Source
\item Original artist unknown: `Dog Caricature'. Unknown Source
\end{enumerate}

\subsubsection{Supplementary Figure 1}
\begin{enumerate}
\item Original artist unknown: `Black Horse Photo'. Unknown Source
\item Marc, Franz: `Blue Horse'. (1911), http://www.franzmarc.com/blue-horse/
\end{enumerate}

\subsubsection{Supplementary Figure 2}
\begin{enumerate}
\item Original artist unknown: `Photo of Woman with Wavy Hair'. \newline http://www.faculty.idc.ac.il/arik/site/foa/face-of-art.asp
\item Leger, Fernand: `La Femme et la fleur'. (1954), https://www.artsy.net/artwork/fernand-leger-la-femme-et-la-fleur
\end{enumerate}

\end{document}